\documentclass[final]{cvpr}
\usepackage{times}
\usepackage{soul}
\usepackage{url}
\usepackage[hidelinks]{hyperref}
\usepackage[utf8]{inputenc}
\usepackage[small]{caption}
\usepackage{graphicx}
\usepackage{amsmath}
\usepackage{amsthm}
\usepackage{booktabs}
\usepackage{algorithm}
\usepackage{algorithmic}
\usepackage{subfig}
\usepackage{color}
\usepackage{comment}
\usepackage{amssymb} 
\usepackage[misc]{ifsym}
\begin{document}

\title{Transformer Guided Geometry Model for Flow-Based Unsupervised Visual Odometry
}


\author{Xiangyu Li,
        Yonghong Hou,
        Pichao Wang,
        Zhimin Gao,
        Mingliang Xu,
        and Wanqing Li
\thanks{Corresponding authors: Zhimin Gao and Pichao Wang}
\thanks{X. Li and Y. Hou are with School of Electronic Information Engineering, Tianjing University, Tianjin, China (e-mail: lixiangyu\_1008@tju.edu.cn; houroy@tju.edu.cn).}
\thanks{P. Wang is with DAMO Academy, Alibaba Group (U.S.), Bellevue, USA (email: pichaowang@gmail.com).}
\thanks{Z. Gao and M. Xu are with School of Information Engineering, Zhengzhou University, Zhengzhou, China (e-mail: iegaozhimin@zzu.edu.cn; iexumingliang@zzu.edu.cn).}
\thanks{W. Li is with Advanced Multimedia Research Lab, University of Wollongong, Wollongong, Australia (email: wanqing@uow.edu.au).}}


\maketitle

\begin{abstract}
Existing unsupervised visual odometry (VO) methods either match pairwise images or integrate the temporal information using recurrent neural networks over a long sequence of images. They are either not accurate, time-consuming in training or error accumulative. In this paper, we propose a method consisting of two camera pose estimators that deal with the information from pairwise images and a short sequence of images respectively. For image sequences, a Transformer-like structure is adopted to build a geometry model over a local temporal window, referred to as Transformer-based Auxiliary Pose Estimator (TAPE). Meanwhile, a Flow-to-Flow Pose Estimator (F2FPE) is proposed to exploit the relationship between pairwise images. The two estimators are constrained through a simple yet effective consistency loss in training. Empirical evaluation has shown that the proposed method outperforms the state-of-the-art unsupervised learning-based methods by a large margin and performs comparably to supervised and traditional ones on the KITTI and Malaga dataset.
\end{abstract}

\section{Introduction}
Visual Odometry (VO) enables agents to infer camera poses (also called Ego-Motion) from image sequences. It has many applications in robotics \cite{robotnavigation,do2010application,mitic2018chaotic}, autonomous driving \cite{autonomousdriving,menze2015object}, augmented/virtual reality \cite{azuma1997survey}, etc. In the past few decades, numerous feature-based or direct photometric visual simultaneous localization and mapping (vSLAM) systems \cite{dso,LSD,orbslam,vinsfusion,VISO} have been proposed and achieved great success in some real-world environment. Some SLAM systems \cite{ding2020digging,li2019a} are also fused with depth information to enhance the system performance. However, these traditional methods often fail in featureless or light-change conditions.

 Recently, deep learning \cite{pwcnet,monodepth} has beem widely applied to many computer vision tasks with promising results. This has led researchers to employ Convolutional Neural Networks (CNNs) or Recurrent Neural Networks (RNNs) in visual odometry. Such supervised learning-based VO \cite{beyond,LSVO,deepvo} requires a large dataset with ground truth of camera poses to train the networks. However, ground truth data is difficult and expensive (e.g. LIDAR and GPS) to collect in practice. Unsupervised methods \cite{sfmlearner,zhan,geonet,ICP,unos} intend to resolve this problem through jointly training the estimation of both depth and camera pose by minimizing the photometric warping loss between forward-backward or left-right image pairs.
 
To date, some unsupervised joint frameworks (e.g. DF-Net \cite{dfnet}, GeoNet \cite{geonet}, Unos \cite{unos}, etc.) take the advantage of geometry characteristic among depth, optical flow and camera pose, and improve the performance through reasoning about rigid and non-rigid (dynamic) scenes. However, they mainly focus on depth and flow, and only pairwise images are taken into consideration, which results heavy error accumulation over time in camera pose task and the results are far lower compared with supervised or traditional methods. Several attempts \cite{beyond,deepvo,wang2019recurrent} have been made to use RNN or Long Short-Term Memory (LSTM) to learn and enforce the temporal correlation over a long period. However, Xue et al. \cite{beyond} has proved that error accumulation cannot be fully eliminated when the historical information stored in single hidden state is directly utilized. In addition, such an approach \cite{wang2019recurrent} is time-consuming in training due to its relying upon the photometric warping loss as supervision.

To address the above problems, an end-to-end self-supervised visual odometry framework is proposed in this work. We decouple the odometry into two branches, one from pairwise images and one from multiple images. First, in the branch for multiple images, a VO is treated as a machine translation-like task in a local temporal window and a full-attentional structure called Transformer \cite{transformer} is adopted as an auxiliary module, referred to as Transformer-based Auxiliary Pose Estimator (TAPE). Depth maps are concatenated with optical flow at an instant in time to form an input (referred to as DF-Group). Multiple DF-Groups are taken as the input `words' and the camera poses at the corresponding times as the output `words'. The geometry correlation among depth, flow and pose and the temporal dependencies of the input DF-Groups are built into TAPE. Second, as features of each DF-Group are mapped to a low dimensional embedding and some information related to pairwise images is inevitably lost. Thus, a Flow-to-Flow Pose Estimator (F2FPE) is proposed by training camera pose estimation from optical flow to exploit both spatial and temporal relationships between image pairs. Third, we propose a loss that enforce the consistency between poses estimated from TAPE and F2FPE. TAPE provides the side information related to temporal correlation and geometry constraint. This side information propagates
back to F2FPE through the pose consistency loss. Thus, F2FPE is taken as the main estimator in the inference stage in our framework.

The main contributions of the paper are summarized as follows: 1) A Transformer-like pose estimator is proposed to effectively provide the geometry and correlation among local multiple frames. This reduces the error accumulation significantly as shown in the experiments, e.g. Fig. \ref{Fig_ablation}. 2) F2FPE takes advantage of the relationship between pairwise images to predict the accurate camera poses. 3) A consistency loss is imposed on the poses estimated from both estimators. The proposed framework is extensively evaluated on KITTI dataset, achieving state-of-the-art performance compared with unsupervised methods and being comparable to supervised and traditional methods. Moreover, our method shows good robust performance on the Malaga urban dataset which contains strong light change scenes.

The remainder of the paper is organized as follows: Section \ref{Section: Related work} provides research outcomes related to the proposed method; Section \ref{Section:Method} describes our architecture and the training scheme;  Section \ref{Section:Experiment} delineates the experimental setting and illustrates the evaluation results with corresponding analysis; Section \ref{Section:Conclusion} offers concluding thoughts and directions for future work.

\begin{figure*}
\begin{center}
\includegraphics[width=0.90\linewidth, height=65mm]{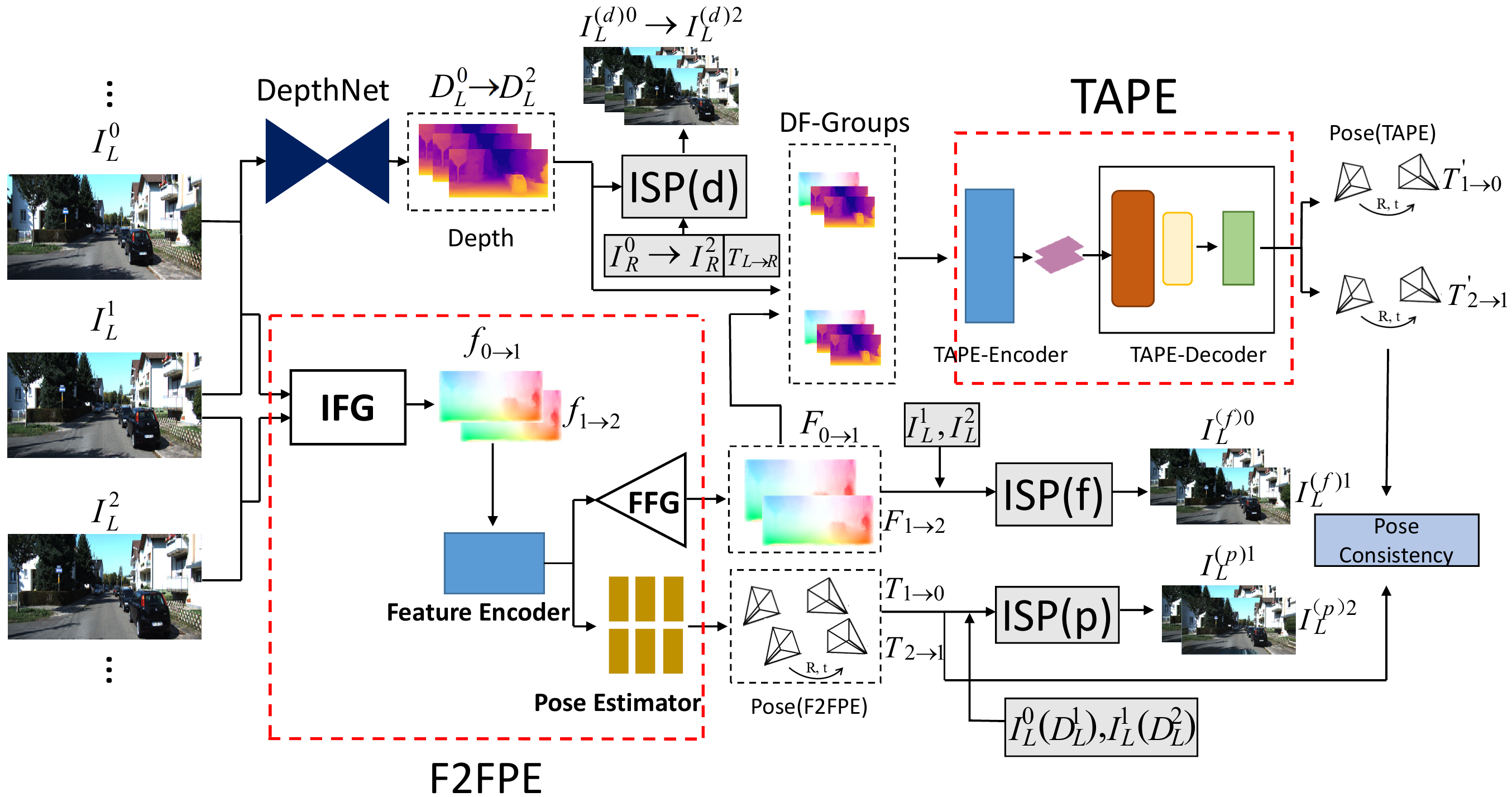}
\end{center}
   \caption{The illustration of the proposed framework, F2FPE takes every two consecutive images as input and predicts corresponding camera poses. TAPE processes multiple DF-Groups (concatenation of depth and optical flow) in parallel to estimate the camera poses in a local temporal window. ISP denotes an image synthesis process, IFG denotes an initial flow generator and FFG represents a final flow generator.}
\label{Fig2}
\end{figure*}

\section{Related works}  \label{Section: Related work}
Visual odometry has been extensively studied in robotics and machine perception for decades, many traditional methods have been proposed and perform effectively in many cases. They can be broadly divided into two categories: feature-based methods \cite{orbslam,PTAM,vinsfusion,VISO} and direct methods \cite{dso,LSD}. Feature-based methods rely heavily on accurate feature association between ima ge pairs, resulting  fragility in region of low texture. Direct methods estimate the ego-motion via minimizing photometric error without calculating the feature points. Since this method is based on the assumption of uniformity of light intensity, it is sensitive to the illumination change and camera exposure parameters. Some biologically inspired methods \cite{stmvo,li2013decentralized,li2013a,hong2018topological} are also proposed to improve the adaptiveness and robustness of VO systems or mapping methods. With recent advances in deep learning, visual odometry has also been attempted with deep learning techniques. This section mainly reviews the related deep learning based methods.

 \textbf{Supervised Methods} \quad Kendall et al. \cite{kendall2015posenet} first proposed a monocular relocalization system, regressing 6-DoF camera poses from raw RGB images using a CNN. Considering dynamics and relation on a sequence of images, DeepVO \cite{deepvo} employs a Recurrent CNN (RCNN) to model VO as a sequential learning problem and achieves comparable results against traditional methods. This method is further improved in \cite{beyond} by introducing a key-state store (Memory) module and a spatial-temporal feature attention (Refine) module. As an important role in VO, optical flow is embedded in several works. LS-VO \cite{LSVO} proposes to jointly train camera pose and flow estimation. In \cite{demon}, depth and camera poses are simultaneously estimated using additional ground truth of optical flow and surface normal. In addition, VINet \cite{VINet} fuses the features from visual and inertial measurement unit (IMU) to train a VO.

\textbf{Unsupervised Methods} \quad Recently, unsupervised methods have attracted increasing attention. Deriving from direct VO methods, photometric warping loss is taken as the supervision in unsupervised VO. Inspired by structure from motion (SfM) techniques, Zhou et al. \cite{sfmlearner}  first  propose to learn depth and camera poses simultaneously from consecutive monocular images. Following the work of Zhou et al. \cite{sfmlearner}, Mahjourian et al. \cite{ICP} construct a 3D geometric loss calculated by Iterative Closest Point (ICP) algorithm. As an extension to the above two methods, optical flow is jointly trained with depth and camera pose in GeoNet \cite{geonet}. In addition, Generative Adversarial Network (GAN) \cite{ganvo} and epipolar geometry constraints \cite{geometric} are adopted to improve the depth and pose estimation. However, camera poses inferred from monocular images suffer from the scale ambiguity. To tackle this problem, in \cite{zhan,undeepvo} stereo sequences with known baseline are taken as input to train their VO systems. Upon the stereo VO, a traditional pose graph back end \cite{posegraph} is utilized to optimize the pose. By leveraging the ConvLSTM, Wang et al. \cite{wang2019recurrent} propose to take arbitrary length sequences as input to deliver a consistent scale. Although recurrent unit are supposed to have the ability to exploit historic information for current pose estimation, Xue et al. \cite{beyond} has proved that only consider historical information stored in a single hidden state could result in severe error accumulation. This paper proposes to build an effective sequential model over a short period using a Transformer like neural network to improve pose estimation.

\textbf{Transformer} \quad RNNs and LSTMs have been widely used in visual odometry by treating VO as a sequence-to-sequence estimation task \cite{deepvo,espvo,beyond}. Transformer \cite{transformer}, a fully-attentional architecture, has achieved better results than RNN or LSTM based methods for sequential modeling problems e.g., machine translation and language modeling. The overall architecture is composed of stacked self-attention and point-wise feed forward layers for both encoder and decoder. Benefiting from the ability of building sequential models, The image transformer \cite{parmar2018image} is proposed for image generation task and Action Transformer~\cite{girdhar2019video} is designed  to recognize and localize human actions in video clips. Inspired by these works, we use a Transformer to exploit and establish the inter-modality geometry dependence among depth, optical flow and camera poses as well as the dependence over a short period.

 For learning-based VO, supervised methods \cite{deepvo,espvo,beyond} exploit RNN or LSTM to model the sequential information, and error is accumulated through the recurrent units; unsupervised VO \cite{zhan,sfmlearner,geonet,ICP,geometric} only take the forward-backward image pairs into consideration, resulting heavy error accumulation on long test sequences. Different from them, this paper proposes two pose estimators, TAPE and F2FPE. TAPE uses Transformer to build the geometry constraint and inter-frame dependencies in a local temporal windows, and F2FPE focuses on image pairs and exploits optical flow to introduce both spatial and temporal relationships. And we propose to enforce the consistency constraint on both estimators so that poses predicted from our framework can leverage the information from pairwise images as well as local temporal window.

\section{Method \label{Section:Method}}
This section begins with an overview of the proposed framework and the supervision signal for self-learning (Sec. \ref{Subsection:Overview}). Then the proposed two modules in the framework, Transformer-based Auxiliary Pose Estimator (TAPE) and Flow-to-Flow Pose Estimator (F2FPE), are described respectively in (Sec. \ref{Subsection:TAPE}) and (Sec. \ref{Subsection:F2FPE}). Finally, we present the loss function and overall network architecture (Sec. \ref{Subsection:Training Losses} and Sec. \ref{Subsection:architecture}).

\subsection{Overview\label{Subsection:Overview}}
Our framework focuses on the camera pose estimation from pairwise images as well as multiple frames in a local temporal window. Fig. \ref{Fig2} shows an overview of the proposed framework. It is composed of a DepthNet and two pose estimators, TAPE and F2FPE. The framework takes a sequence of left or right images, e.g. three consecutive images as illustrated in the figure, as input and the estimated relative camera poses as output, wherein depth and optical flow are estimated for image synthesis to self-supervise the training. In order to recover the absolute scale of camera poses, the DepthNet is trained using stereoscopic image pairs similar to \cite{zhan,undeepvo}. The F2FPE module predicts poses and flow between adjacent two images. Furthermore, TAPE estimates poses among multiple DF-Groups over a short period, e.g. three frames/images. A DF-Group consists of depth maps for two adjacent images and one optical flow of the two images. In this paper, TAPE is evaluated on two DF-Groups. Camera poses predicted from both F2FPE and TAPE are constrained by the consistency loss.

 Without loss of generality, as shown in Fig. \ref{Fig2}, let the consecutive image pairs to the F2FPE module be ($I_L^0$, $I_L^1$), where $L$ indicates the left view and $0$, $1$ denote two consecutive time stamps, output camera pose be $T_{1\rightarrow 0}$ and optical flow be $F_{0\rightarrow 1}$. $D_L^0$ and $D_L^1$ are the depth maps of $I_L^0$ and $I_L^1$ estimated by the DepthNet and $T_{L \rightarrow R}$ is known stereo camera pose. Given image pairs ($I_L^0$, $I_L^1$) ($I_L^0$, $I_R^0$), we can project pixel coordinates from one view onto corresponding view according to \cite{sfmlearner} and then images $I_L^{(p)1}$, $I_L^{(d)0}$ and $I_L^{(f)0}$ can be synthesized using differential bi-linear interpolation \cite{spatial}. \{$p, d, f$\} denotes the image synthesis for camera pose, stereo depth and optical flow respectively. Computing the photometric error between input images and the synthesized ones forms a fundamental training loss in our framework. This will be described in Sec. \ref{Subsection:Training Losses}.

\subsection{TAPE \label{Subsection:TAPE}}
TAPE, a Transformer-style pose estimator, is built to model the geometry and temporal information over a short period. TAPE treats the pose estimation as a machine translation task and predicts camera poses from a few DF-Groups, each DF-Group being a concatenation of optical flow and depth maps instead of RGB images. Such a design is based on the theoretical results in \cite{roberts2014optical} that there is dependence among the camera pose, depth and optical flow. TAPE is a nonlinear approximator of the geometry dependence.

\begin{figure}[!t]
\begin{center}
\includegraphics[width=0.85\linewidth, height = 40mm]{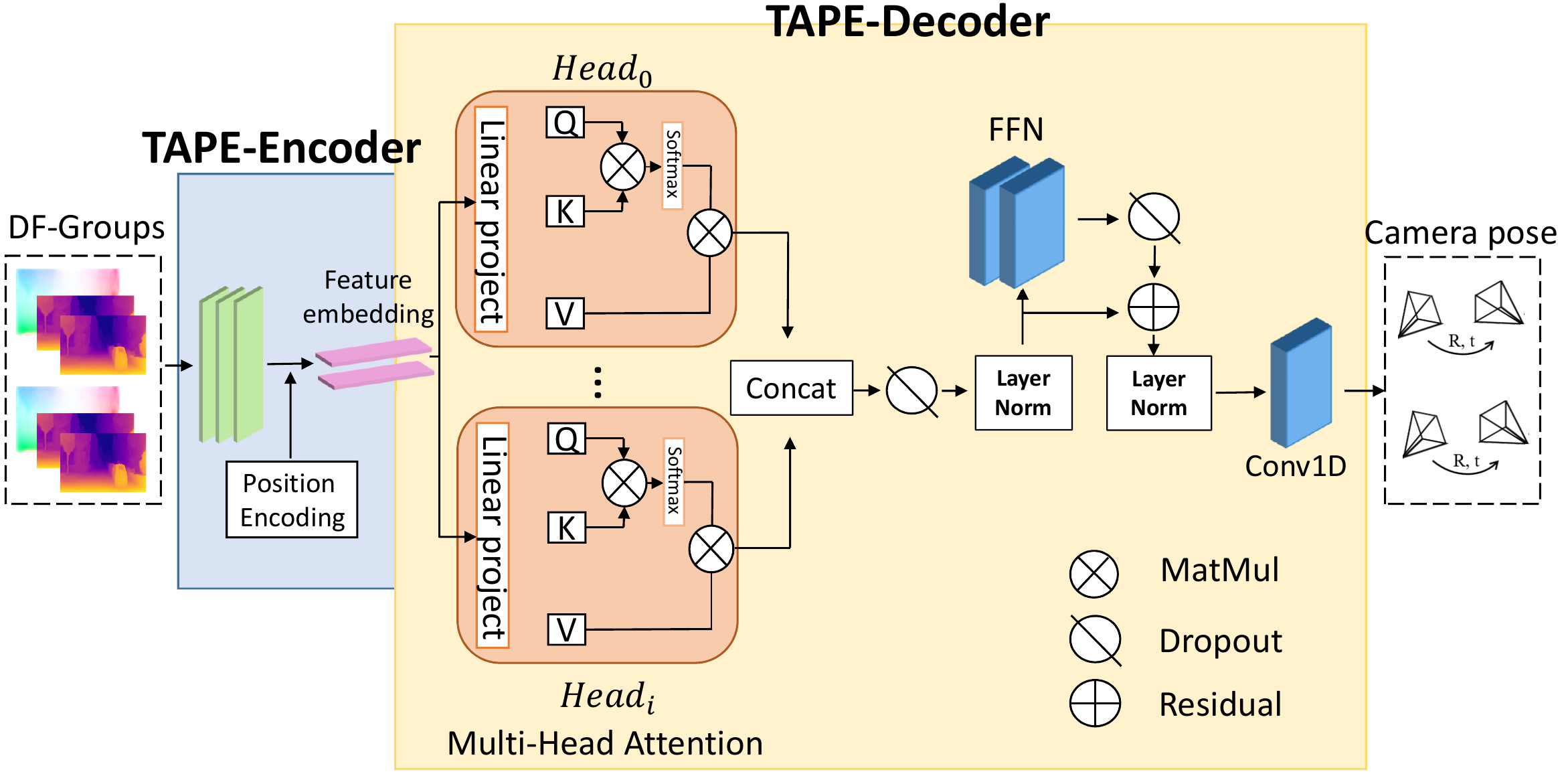}
\end{center}
   \caption{The architecture of TAPE. DF-Groups are fed to TAPE-Encoder to obtain the feature embedding, then self-attention mechanism is applied to the linear projections of the embedding. The concatenation of the output from each head passes through a series operations to obtain the final camera poses. Notice that FFN stands for a Feed-Forward Network.}
\label{Fig3}
\end{figure}

Specifically, TAPE takes multiple DF-Groups and predicts corresponding poses, e.g., ($D_{L}^0$, $D_{L}^1$, $F_{0 \rightarrow 1}$) and ($D_{L}^1$, $D_{L}^2$, $F_{1 \rightarrow 2}$) as input and $T'_{1 \rightarrow 0}$, $T'_{2 \rightarrow 1}$ as the output. Therefore, TAPE translates DF-Groups (`input words') to camera poses (`output words') in an one-to-one correspondence. As shown in Fig. \ref{Fig3}, TAPE contains two stages, TAPE-Encoder and TAPE-Decoder. The first stage encodes the DF-groups into features, then directly applies average pooling across space to obtain the feature embeddings with size of $d_{model}$. Information about relative positions of feature embedding is injected using position encoding. After feature embedding, the TAPE-Decoder first maps features to a series of low dimensional vectors query ($Q$), key ($K$) and value ($V$) with dimensions of ($d_k, d_k, d_v$) using learnable linear projection. Then the multi-head attention is applied to these vectors. For each head, the output can be represented by
\begin{equation}
{\rm Head}_i = {\rm Softmax}(\frac{QK^T}{\sqrt{d_k}})V
\end{equation}
where the dot-product of $Q$ and $K$ vectors are normalized by $\sqrt{d_k}$ in order  to prevent the softmax function from suffering from small gradients. Followed by a group of operations including dropout, residual connection and LayerNorm (LN) \cite{ba2016layer}, the final results can be obtained by
\begin{gather} 
{\rm q'} = {\rm LN}(F+{\rm Dropout}({\rm Concat}({\rm Head}_i)))\\ 
{\rm {T'}} = {\rm Conv1D}({\rm LN}({\rm q'}+{\rm Dropout}({\rm FFN}({\rm q'})))
\end{gather}
where $F$ is the matrices packing of input feature embeddings, $\rm q'$ is the intermediate query feature during processing and the Feed-Forward Networks (FFN) is composed of two convolution layers with kernel size being $1$.

 Camera poses predicted directly from a RGB paired images without geometry information is prone to error. By implanting Transformer in VO, TAPE models the inherent geometry correlation between multiple DF-Groups and corresponding camera poses and improve the performance. Meanwhile, compared with recurrent units, our TAPE can process the DF-Groups in parallel to utilize the temporal correlation better which will be verified in Sec. \ref{Ablation studies}. In TAPE, each DF-Group represents a RGB image pairs. Due to the average pooling in feature embedding, some internal information between pairwise images is lost. Therefore, another estimator is needed to enhance the pose estimation from image pairs.

\subsection{F2FPE  \label{Subsection:F2FPE}}
The Flow-to-Flow Pose Estimator (F2FPE) is proposed to focus on the pose regression between two consecutive frames. As shown in Fig. \ref{Fig2}, F2FPE contains four parts: Initial Flow Generator (IFG), Feature Encoder (FE), Pose Estimator (PE) and Final Flow Generator (FFG). IFG usually takes a pre-trained flow network to produce an initial optical flow, then the FE and FFG form a encoder-decoder architecture to refine the optical flow. The shared features are fed to the fully-connected PE to predict the camera poses. The process can be expressed as

\begin{gather}
{f_{0\rightarrow 1}} = {\rm IFG}(I_L^0, I_L^1)\\
{F_{0\rightarrow 1}}={\rm FFG}( {\rm Encode}( f_{0\rightarrow 1}))\\
{T_{1\rightarrow 0}}= {\rm PE}( {\rm Encode}( f_{0\rightarrow 1}))
\end{gather} where the $(I_L^0$, $I_L^1)$ are stacked along RGB channels and features extracted from initial optical flow $f_{0\rightarrow 1}$ are shared by pose-branch and final-flow branch.

Since VO predicts the relative camera pose from the reference frame to current frame, it relies on the relationship between pairwise images. Several works \cite{Deeptam,LSVO,deepvo} have proved that features extracted from optical flow are fit for frame-to-frame pose estimation because of its representation of pixel motion relationship between two frames. Thus, optical flow has been employed as either input or output in visual odometry. We also design our F2FPE based on optical flow, and it has two advantages. First, both flow from IFG or FFG can be trained in an unsupervised manner instead of using ground truth \cite{Deeptam,LSVO}. Second, F2FPE can be flexibly adapted to different scenarios. In the case of virtual scenes where optical flow ground truth are available, the IFG can be trained in a supervised manner to produce optical flow that is sufficiently accurate for pose estimation. In this case, FFG can be removed from F2FPE. In the case of real-world scenes, e.g. KITTI \cite{kitti} or Malaga \cite{malaga} dataset, the little or no ground truth is available, FFG refines the optical flow from an initial flow generated by IFG.

\subsection{Loss for Training  \label{Subsection:Training Losses}}
The loss for training consists of several components as described below.

\textbf{Image Synthesis Loss} \quad As mentioned in Sec. \ref{Subsection:Overview}, image synthesis loss, denoted as $L_{is}$, acts as the main loss to train the proposed framework in an unsupervised manner.  $L_{is}$ is calculated between a reconstructed image $I^{(*)}$ and its original one $I$ . Combing L1 norm and SSIM (structural similarity index) \cite{ssim}, $L_{is}$ can be expressed as
\begin{equation}
L_{is}=\sum_{* \in p, d, f}(\alpha\frac{1-SSIM(I^{(*)},I)}{2}+(1-\alpha){||I^{(*)}-I||_1} \label{loss_ap}),
\end{equation}
where $\alpha$ is a weight balancing structure and appearance. Note that for every pair of images, reconstruction can be bi-directional i.e. forward-to-backward and backward-to-forward, left-to-right and right-to-left.

 \textbf{Depth Loss} \quad Although our framework does not specifically optimize depth estimation, accurate depth estimation contributes significantly to pose estimation. Therefore, a few depth losses are introduced in the training. Similar to monodepth \cite{monodepth}, a left-right depth consistency loss $L_{lr}$, an edge-aware smooth term $L_{sm}$ and additional regularization term $L_{reg}$  are employed in order to produce  accurate and locally smooth depth maps that are used in the framework.

 \textbf{Pose Consistency Constraint} \quad We propose a simple yet effective loss function that enforces the consistency between the two estimators. Thus the poses predicted from our framework can leverage the information from pairwise images as well as local temporal window.
\begin{equation}
L_{pc}=\sum_{N}||T'-T||_1
\end{equation}
where $T'$ and $T$ denotes the poses from TAPE and F2FPE respectively, $N$ is the number of relative poses between multiple frames e.g., $(N+1)$ frames can predict $N$ camera poses.

 \textbf{Overall loss} \quad The overall loss function in our framework can be expressed by
\begin{equation}
L_{all}=L_{is}+\lambda_{pc} L_{pc}+ \lambda_{sm} L_{sm}+\lambda_{lr} L_{lr}+\lambda_{reg} L_{reg} 
\end{equation}
where $\lambda_{pc}$, $\lambda_{sm}$, $\lambda_{lr}$ and $\lambda_{reg}$ are weights for the loss terms.

\subsection{Network Architecture  \label{Subsection:architecture}}
Both TAPE and F2FPE are plug-and-play modules and they can be easily integrated into other unsupervised VO systems. The basic network configurations adopted for experimental verification of the proposed framework is introduced below.

 \textbf{F2FPE} \quad In order to produce appropriate initial optical flow for the feature encoder, PWC-Net \cite{pwcnet} is adopted as the IFG. Feature Encoder and FFG form a fully convolutional encoder-decoder structure. The flow feature extracted by Feature Encoder with seven stride-2 convolution layers and then enlarged to four scales optical flow by FFG with seven deconvolution layers. Meanwhile, followed by two separate groups of three fully-connected layers, a 6-DoF camera pose consisting of three Euler angles and 3-D translation is regressed.

 \textbf{TAPE} \quad ResNet-50 \cite{resnet} is adopted as the backbone of TAPE-Encoder and followed by average pooling to produce the feature embedding. The size of feature embedding $d_{model}$ = 512 and the number of heads is set to 2. In TAPE-Decoder, each size of group ($Q$, $K$, $V$) are set to ($d_k$ = $d_v$ = $d_{model}$/2 = 256). Dropout rate 0.1 is used in the framework.

 The DepthNet is mainly based on monodepth \cite{monodepth} with a replacement of the disparity output with inverse depth. Note that the input to the DepthNet is left images, and the right images are used only for image synthesis.

\section{Experiments \label{Section:Experiment}}

\subsection{Implementation and Training  \label{Subsection:Implementation}}

The framework is implemented in Tensorflow \cite{tensorflow}. Adam optimizer with ($\beta_1$=0.9, $\beta_2$=0.999) is used to train all networks. The initial learning rate is set to 0.0002 and goes down by half for every 1/5 of the total iterations. The weights are set to [$\lambda_{sm}$, $\lambda_{lr}$, $\lambda_{reg}$, $\lambda_{pc}$] = [0.1, 0.4, 0.02, 1.0 (10.0)], $\lambda_{pc}$ is set to 10.0 in baseline method (only contains Feature Encoder (FE) and Pose Estimator (PE) in our framework) and 1.0 for full framework because the loss in former is low compared to other losses in our framework and $\alpha$ = 0.85 for training losses.  $\lambda_{sm}$, $\lambda_{lr}$, $\lambda_{reg}$ and $\alpha$ are used for depth training and image synthesis loss, and we mainly follow Godard et al.'s setup~\cite{monodepth}. The parametric study of $\lambda_{pc}$ are reported in Table \ref{table:ablation2}. Since image warp is a memory-consuming process, the framework is trained on a single Titan X GPU with the sequence length of input to TAPE set to 3, and 2 GPUs with the sequence length set to 5. The batch-size is set to $4$ and we train the framework for around 40 epochs. The image size fed to the networks is resized to $256\times512$, and data augmentation with brightness, gamma and color shift is applied to input data randomly. In order to train the framework more effectively, we first train DepthNet separately until it is converged. For the F2FP2, since the training of IFG is time-consuming, the IFG module in our framework is initialized with the pre-trained flow network weight (on FlyingThings3D \cite{flying3d} and FlyingChairs \cite{flyingchairs}) for possibly accelerating the training.

\subsection{Dataset  \label{Subsection:Dataset}}
 \textbf{KITTI} \quad For fair comparison with existing methods, the widely used official KITTI Odometry Split \cite{kitti} for visual odometry benchmark is used. The odometry dataset contains 11 driving sequences with available ground truth of camera poses. Following the same recommended train/test set split for unsupervised and supervised VO, we take corresponding sequences for training and evaluating, i.e. 00-08 sequences are used as the training set in unsupervised comparison while sequences 00, 01, 02, 08, 09 are used to train model for comparison with supervised methods. All the training data are cut to 3-frame or 5-frame segments as input to networks. For depth estimation, the Eigen split \cite{eigen} is adopted to evaluate the DepthNet where 23488 frames are used for training and 697 frames for testing.

\textbf{Malaga} \quad To evaluate the robustness of the proposed framework, Malaga urban dataset \cite{malaga} is used. This dataset has different illumination conditions and camera parameters compared with the KITTI dataset. Sequences 01, 02, 03, 04, 09, 11, 15 are used for training which contains 8928 images and two challenge sequences 08 and 14 are used for evaluation.

\begin{table}[t]
\caption{Quantitative comparison of visual odometry with unsupervised methods on sequences 09, 10. $t_{err}(\%)$ is average translational error and $r_{err}$ is average rotational error. (*) indicates the IFG is also trained in our framework.}
\begin{center}
\begin{scriptsize}
\setlength{\tabcolsep}{0.5mm}{
\begin{tabular}{|l||c|c|c|c|}
\hline
{Method} & \multicolumn{2}{|c|}{Seq.09} & \multicolumn{2}{|c|}{Seq.10}  \\
\cline{2-5}
{ }  & {$\ t_{err}(\%)$}  & {$r_{err}(^\circ/100m)$}    & {$t_{err}(\%)$}   & {$r_{err}(^\circ/100m)$}\\
\hline
Zhou et al. \cite{sfmlearner}    &{17.84}	&{6.78}	&{ 37.91}	&{17.8} \\
\hline
GeoNet \cite{geonet}    &{43.76}	&{16.00}	&{ 35.60}	&{13.8} \\
\hline
Zhan et al. \cite{zhan}           &  11.93&	3.91&	12.45&	3.46\\
\hline
Wang et al.  \cite{wang2019recurrent}  & 9.88  & 3.40  & 12.24  & 5.2       \\
\hline
Li et al. \cite{posegraph}    &8.10	&2.81 &	12.9 &	3.17   \\
\hline
Li et al. (LC) \cite{posegraph}    &6.23	&2.11 &	/ &	/    \\
\hline
UnOS \cite{unos}    &{5.21}	&{1.80}	&{ 5.20}	&{2.18} \\
\hline
{\bfseries Ours (TAPE)}   &{ 6.72}	&{ 2.60}	&{  8.66}	&{ 3.13}\\
\hline
{\bfseries Ours (F2FPE)}   &{ 2.79}	&{ 1.07}	&{  4.16}	&{ 1.32}\\
\hline
{\bfseries Ours* (F2FPE)}    &{\bfseries 2.36}	&{\bfseries 1.06}	&{\bfseries 3.00}	&{\bfseries 1.28} \\
\hline
\end{tabular}}
\end{scriptsize}
\footnotesize
\end{center}
\label{table1}
\end{table}

\begin{figure}[!t]
\centering
\subfloat[Translation against path length]{\label{Fig5a}\includegraphics[width = 0.45\columnwidth]{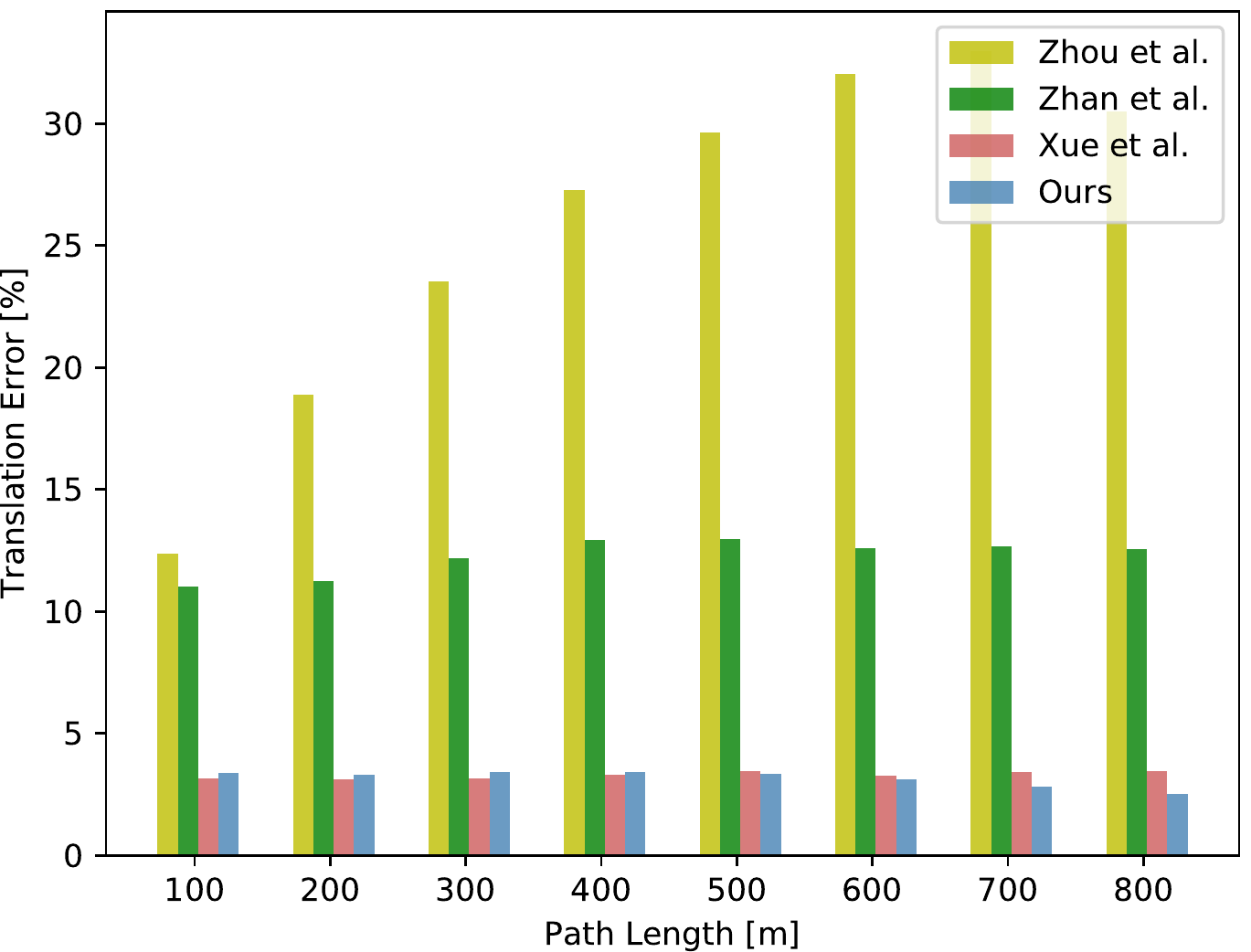}}\
\subfloat[Rotation against path length]{\label{Fig5b}\includegraphics[width = 0.45\columnwidth]{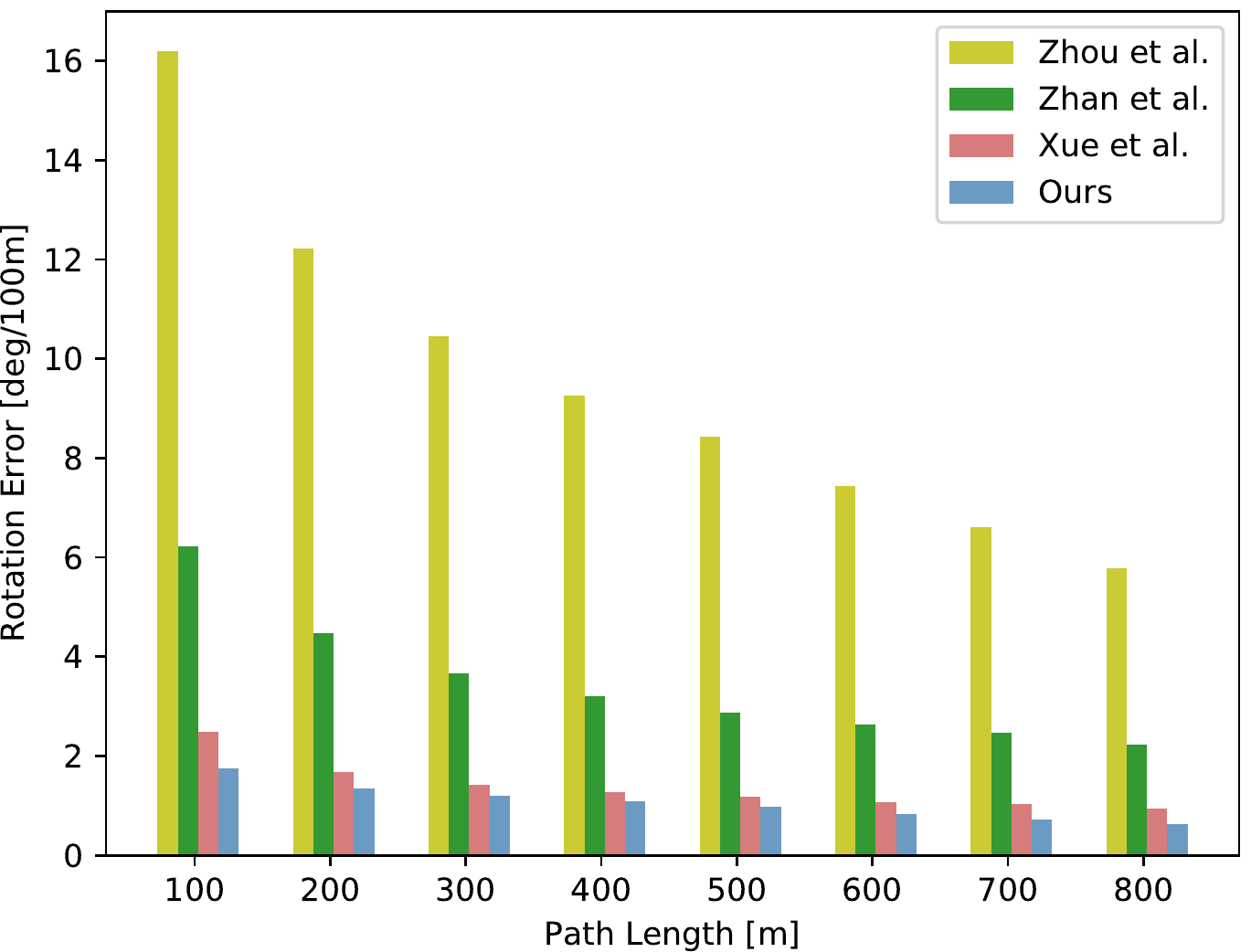}}\\
\subfloat[Translation against speed]{\label{Fig5c}\includegraphics[width = 0.45\columnwidth]{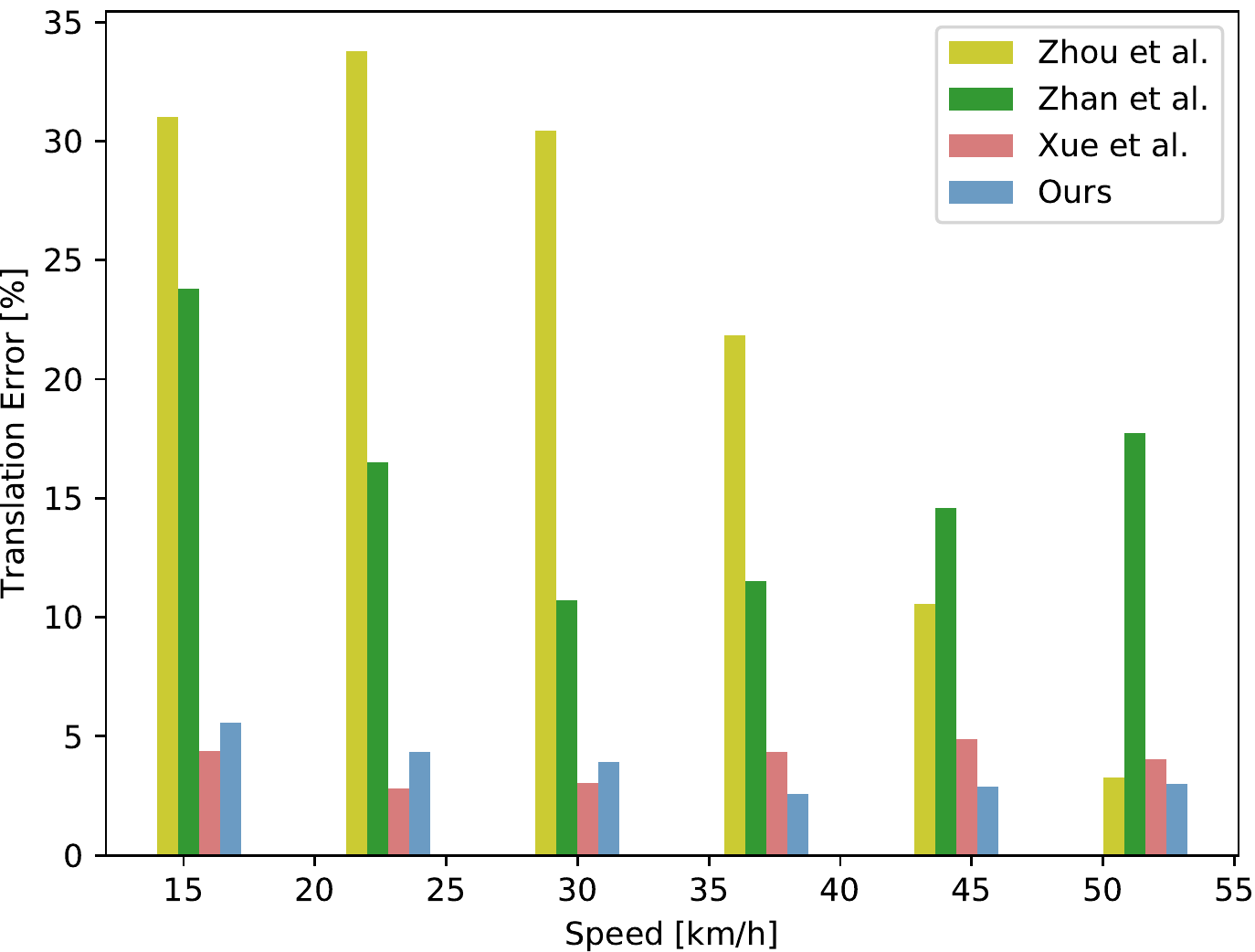}}\
\subfloat[Rotation against speed]{\label{Fig5d}\includegraphics[width = 0.45\columnwidth]{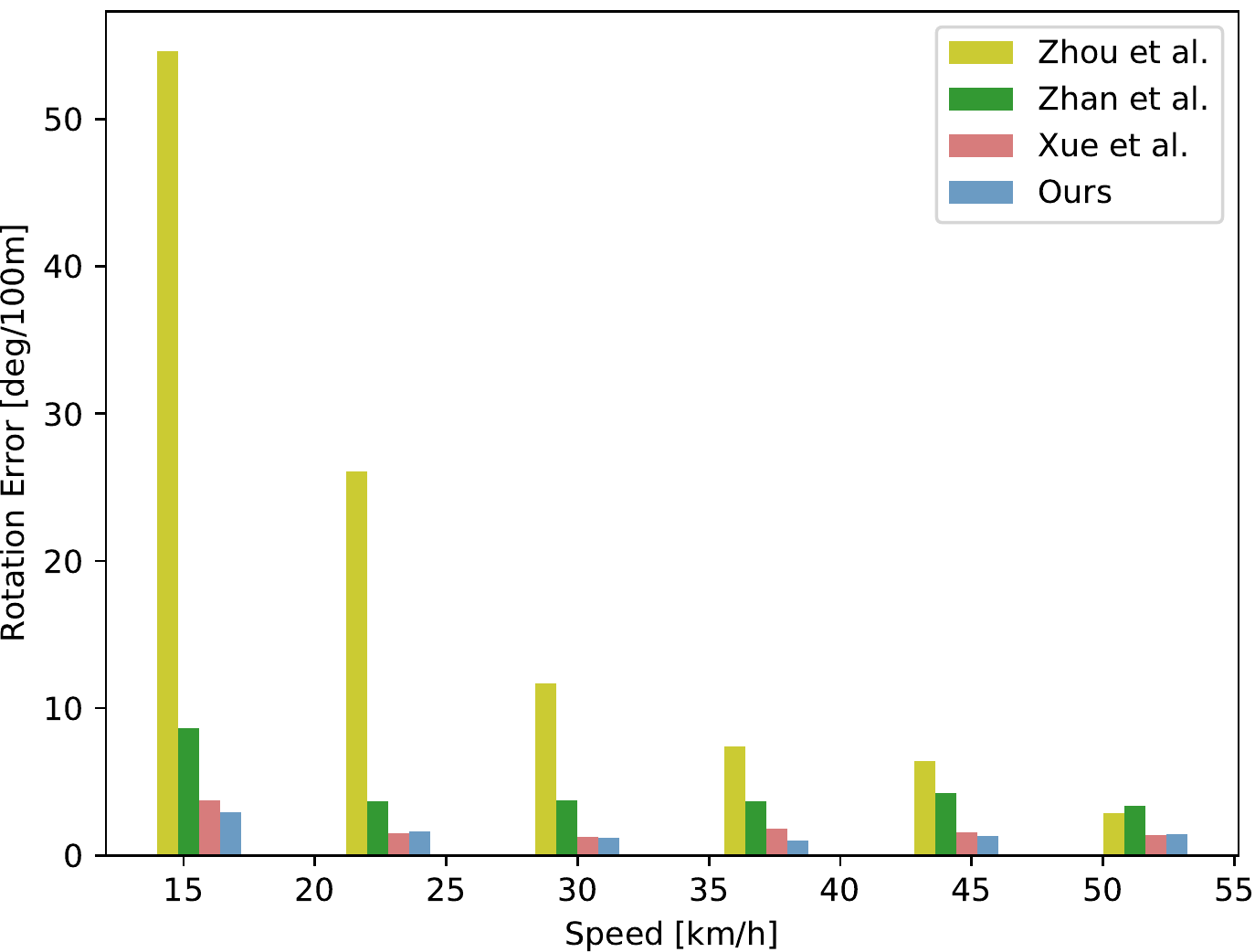}}\
\caption{Average errors on translation and rotation against different path lengths and speeds.}
\label{Fig5}
\end{figure}

\subsection{Visual Odometry  \label{Subsection:visual odometry}}
\subsubsection{KITTI \label{Subsubsection:KITTI}} 
The performance of visual odometry is evaluated using KITTI Odometry metric, which all possible sub-sequences of length (100, 200, ..., 800) meters are used to compute the translational and rotational errors. As shown in Table~\ref{table1}, the results of our framework are compared to the SOTA unsupervised learning-based VO \cite{zhan,wang2019recurrent,sfmlearner,geonet,unos,posegraph}.

\begin{table*}[ht]
    \caption{Results of Seq.03, 04, 05, 06, 07, 10 on KITTI Odometry split. DeepVO \cite{deepvo}, ESP-VO \cite{espvo} and \cite{beyond} are supervised methods. VISO2-M \cite{VISO} and ORB-SLAM2 (with and without loop closure) \cite{orbslam} are traditional methods. Our method is trained on 00, 01, 02, 08, 09.}
\begin{center}
\begin{small}
    \begin{tabular}{|l|c|c|c|c|c|c|}
    \hline
    {Method}  & {Seq.03}& {Seq.04}&{Seq.05}&{Seq.06} & {Seq.07}& {Seq.10} \\

    { }  & $t_{err}$\quad$r_{err}$& $t_{err}$\quad$r_{err}$&$t_{err}$\quad$r_{err}$ & $t_{err}$\quad$r_{err}$& $t_{err}$\quad$r_{err}$ &$t_{err}$\quad$r_{err}$ \\
    \hline
    VISO2-M \cite{VISO}  & 8.47\quad8.82 &4.69\quad4.49 &	19.22\quad17.58	&7.30\quad6.14 & 23.61\quad 29.11 &	41.56\quad32.99	\\
    \hline
    ORB-SLAM2 \cite{orbslam} & 0.97\quad0.19 & {\bfseries 1.30}\quad{\bfseries 0.27} &	9.04\quad0.26	&14.56\quad0.26 & 9.77\quad 0.34 &	2.57\quad0.32 \\
    \hline
    ORB-SLAM2(LC) \cite{orbslam}  & {\bfseries 0.91}\quad {\bfseries 0.19} &1.56\quad0.27 &	{\bfseries 1.84}\quad {\bfseries 0.20}	&4.99\quad0.23 & {\bfseries 1.91} \quad {\bfseries 0.28} &	{\bfseries 3.30}\quad {\bfseries 0.30}\\
    \hline
    \hline
    DeepVO \cite{deepvo} & 8.49\quad6.89 &	7.19\quad6.97 &	2.62\quad3.61	&5.42\quad5.82 & 3.91 \quad 4.60 &	8.11\quad8.83	 \\
    \hline
    ESP-VO \cite{espvo}  & 6.72\quad6.46 &6.33\quad6.08 &	3.35\quad4.93	&7.24\quad7.29 & 3.52 \quad 5.02 &	9.77\quad10.2 \\
    \hline
     Xue et al. \cite{beyond} & 3.32\quad2.10 &	2.96\quad {\bfseries 1.76} & {\bfseries 2.59}\quad{\bfseries 1.25}	& 4.93\quad1.90 & {\bfseries 3.07} \quad {\bfseries 1.76} &	3.94\quad {\bfseries 1.72} \\
    \hline
    {\bfseries Ours} &{\bfseries 2.35}\quad {\bfseries 1.55} & {\bfseries 2.46}\quad1.95 &	3.11\quad1.50	& {\bfseries 2.56} \quad {\bfseries 1.02}  & 7.57 \quad 4.64 &	{\bfseries 3.86}\quad1.82\\
    \hline
    \end{tabular}
    \end{small}
    \end{center}
    \label{table2}
\end{table*}

 Since monocular VO methods Zhou et al. \cite{sfmlearner} and GeoNet \cite{geonet} suffer from scale ambiguity, the results of which are aligned with ground truth for evaluating. Similar to stereo-based VO \cite{zhan,unos,posegraph}, we only integrate the poses predicted from every two frames over the whole sequence without post-processing. As shown in Table \ref{table1}, our method outperforms other unsupervised methods by a large margin. For the monocular VO \cite{sfmlearner,geonet}, the error mainly caused by scale ambiguity. Wang et al. \cite{wang2019recurrent} adopt the ConvLSTM to integrate the temporal information over a long sequence, however the error is also accumulated through recurrent unit. Both Zhan et al. \cite{zhan} and UnOS \cite{unos} use the stereo pairwise images to train the pose estimator, and UnOS proposes to find the rigid area to refine camera poses, however the temporal correlation is not taken into consideration. Li et al. \cite{posegraph} also predicts camera poses from a local temporal window and loop closure is deployed to reduce error drift. However, our method can produce more accurate results. Note that the results of TAPE are inferior to F2FPE, since the inputs of TAPE are estimated from RGB images which also serve as input to F2FPE. On the other hand, F2FPE is trained with the side geometry and temporal information through TAPE, F2FPE alone becomes a natural choice for inference and this would also reduce the inference time by not involving TAPE in inference.

 Fig. \ref{Fig5} shows the average translation and rotation errors over sub-sequences of length 100 up to 800 meters and different vehicle speeds. It can be seen that as the length of trajectory increases and the speed goes fast, both errors of our method are lower than those of other methods. Supervised VO \cite{beyond} proposes a refine module to mitigate the error accumulation over a long sequence, while our method produces lower error due to the collaborative training of TAPE and F2FPE.

\begin{figure}[!t]
\centering
\subfloat[Seq.03]{\includegraphics[width = 0.49\columnwidth, height = 30mm]{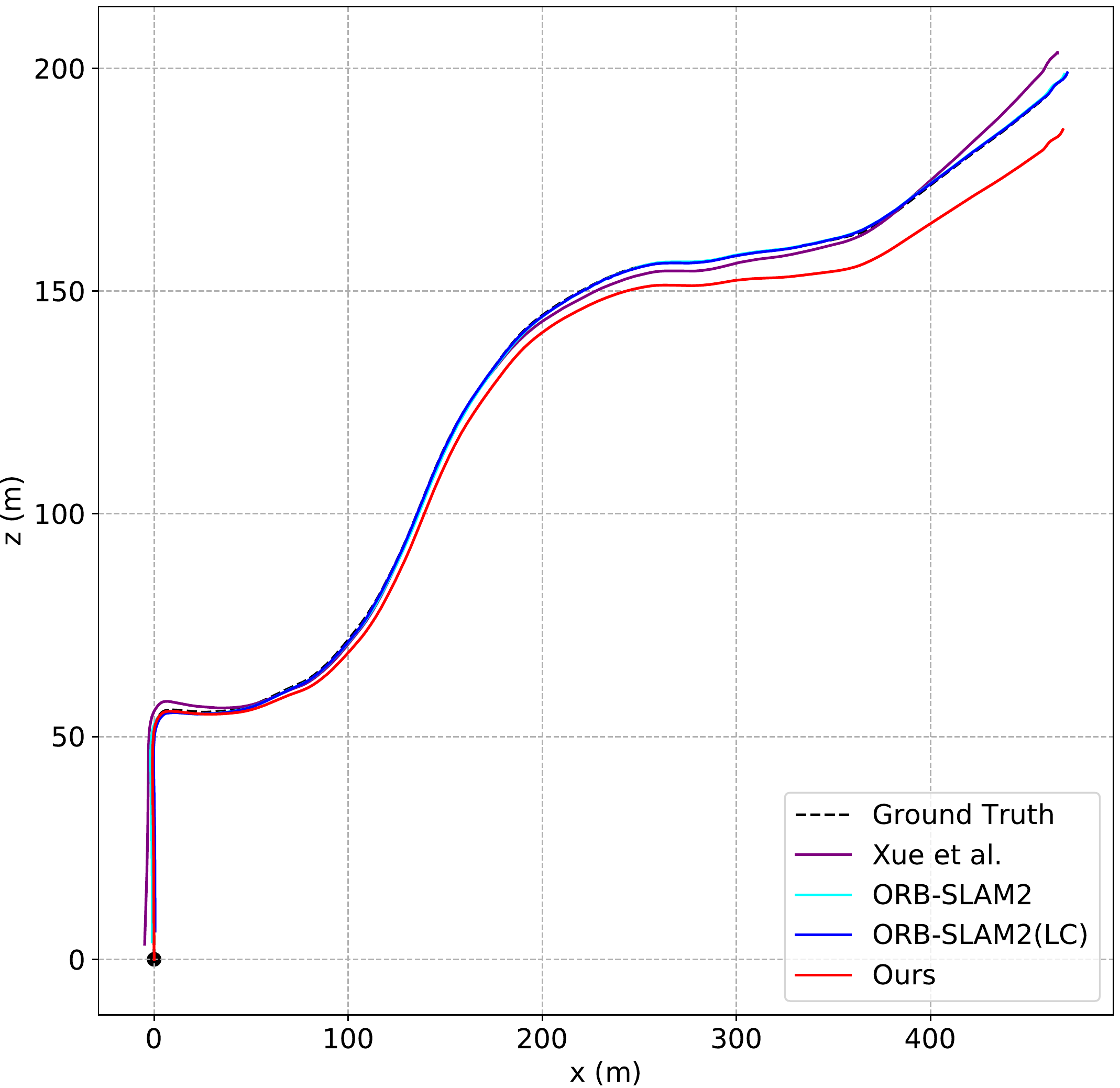}}
\subfloat[Seq.06]{\includegraphics[width = 0.47\columnwidth, height = 30mm]{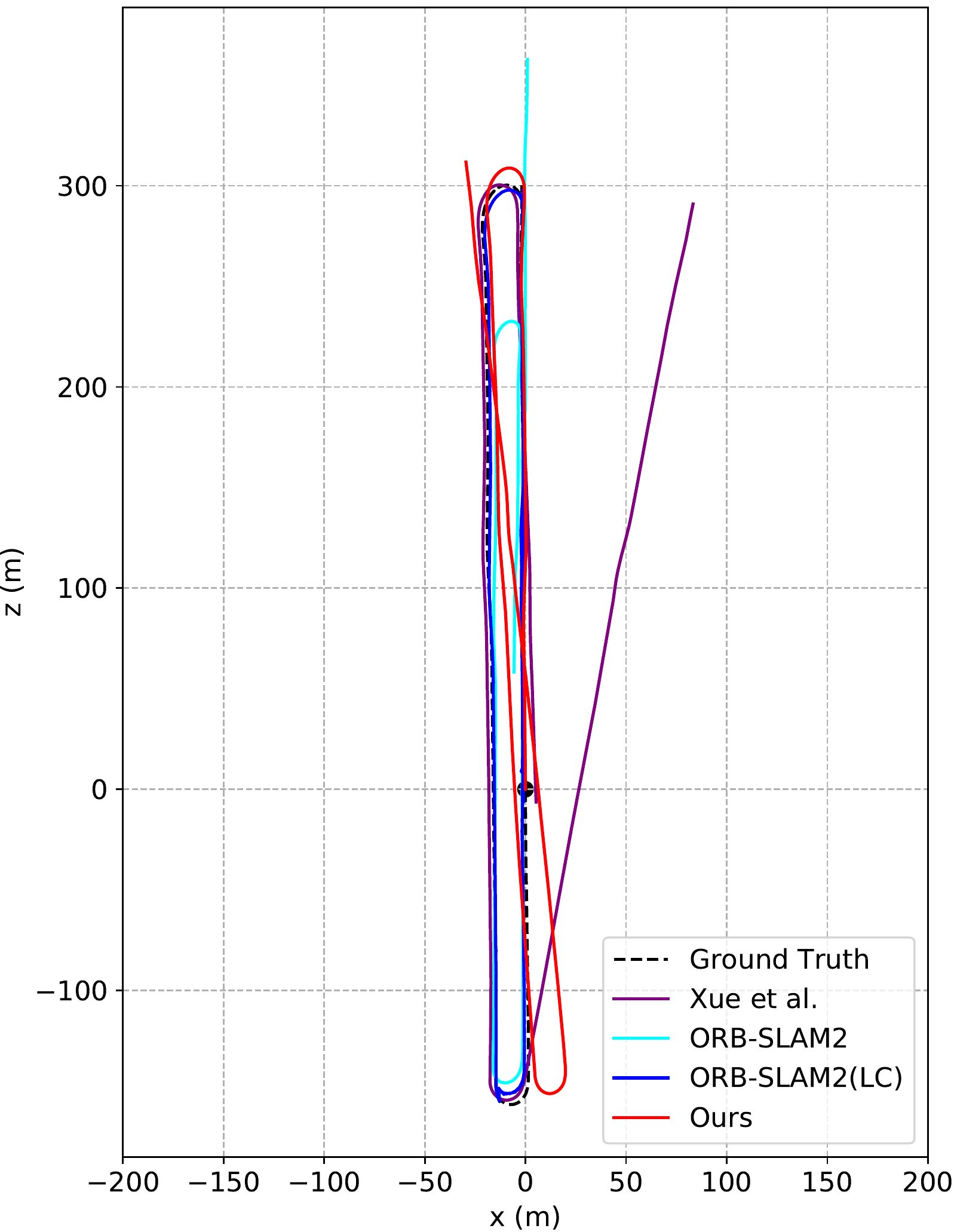}}\
\subfloat[Seq.09]{\includegraphics[width = 0.44\columnwidth, height = 38mm]{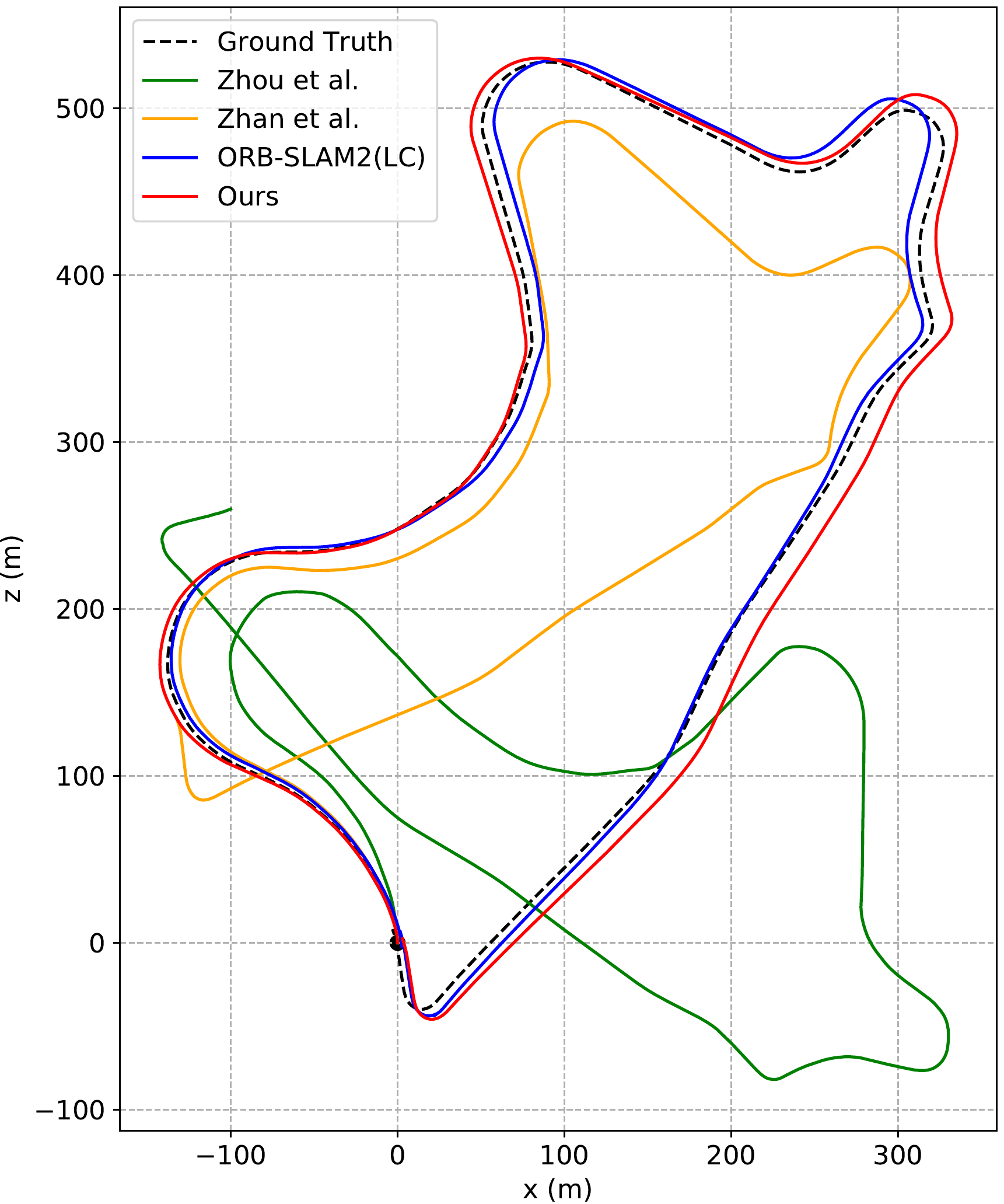}}\quad
\subfloat[Seq.10]{\includegraphics[width = 0.44\columnwidth, height = 38mm]{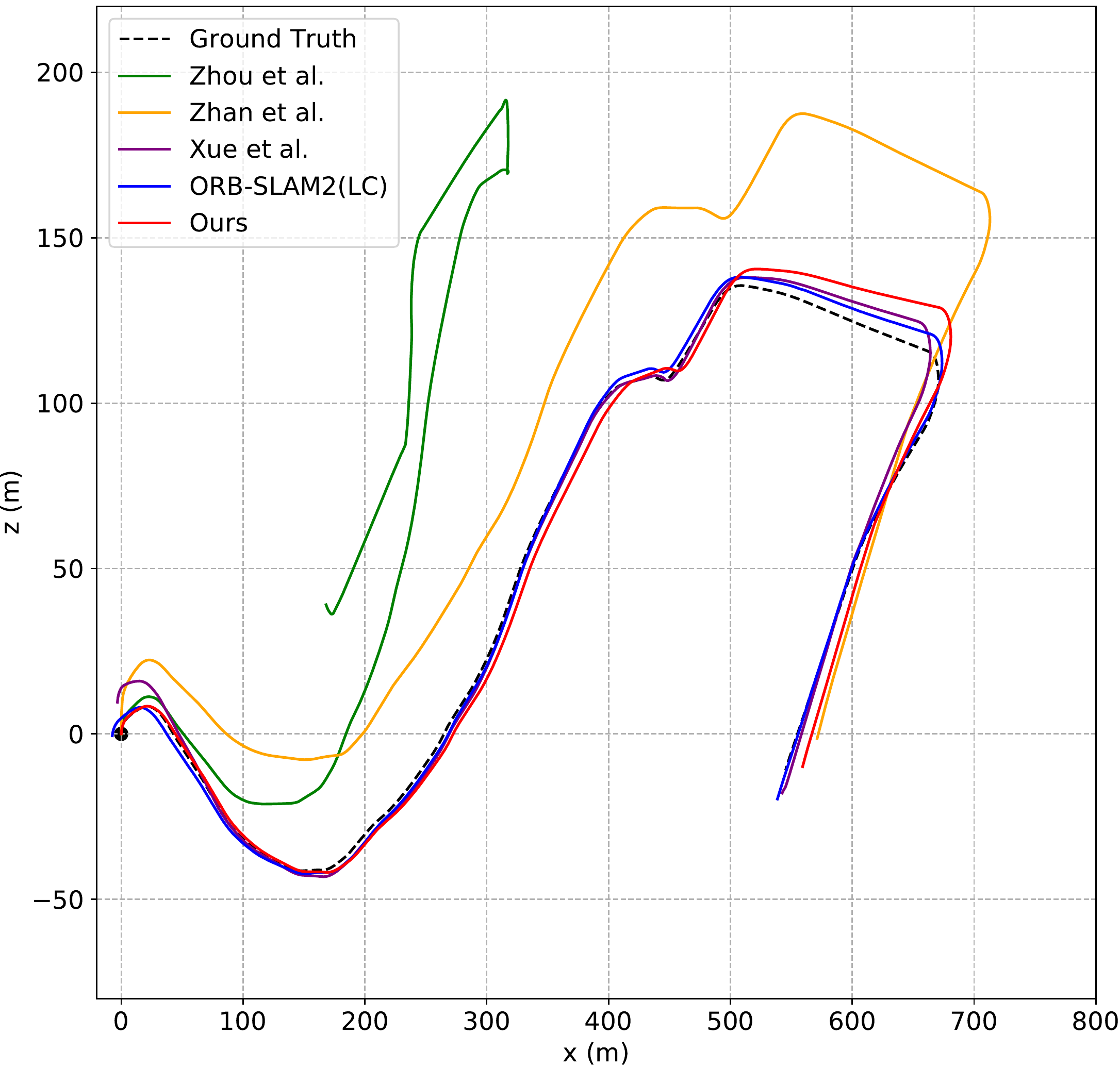}}
\caption{Trajectories of different methods on Seq. 03, 06, 09 and 10.}
\label{Fig6}
\end{figure}

 In addition, our framework is compared with supervised methods \cite{deepvo,espvo,beyond} and traditional VO systems VISO2-M \cite{VISO} and ORB-SLAM2 (with and without loop closure) \cite{orbslam}. Sequence 00, 01, 02, 08, 09 are used for training while the rest for evaluating. As shown in Table \ref{table2}, our method performes better than DeepVO \cite{deepvo}, ESP-VO \cite{espvo}, VISO2-M \cite{VISO} and being comparable to Xue et al. \cite{beyond} in most of the sequences. Xue et al. \cite{beyond} and ORB-SLAM2 \cite{orbslam} respectively utilize a refining module and local bundle adjustment (global loop closure) to optimize the poses. For Seq. 07, our method performs poorly since there are almost 60 static frames. For moving vehicles, it will stop for waiting the traffic light or other cars and pedestrian pass, thus the camera should be in the static status. However, the context of each frames is changing by the time so that our method will still predict the wrong poses in static period. This is the disadvantage of unsupervised VO since the supervision is based on the photometric loss. Such an issue could be mitigated by the bundle adjustment or increasing the training data. Fig. \ref{Fig6} shows visual comparisons of full trajectories.

\begin{figure}[!t]
\centering
\subfloat[Seq.08]{\includegraphics[width = 0.58\columnwidth, height = 54mm]{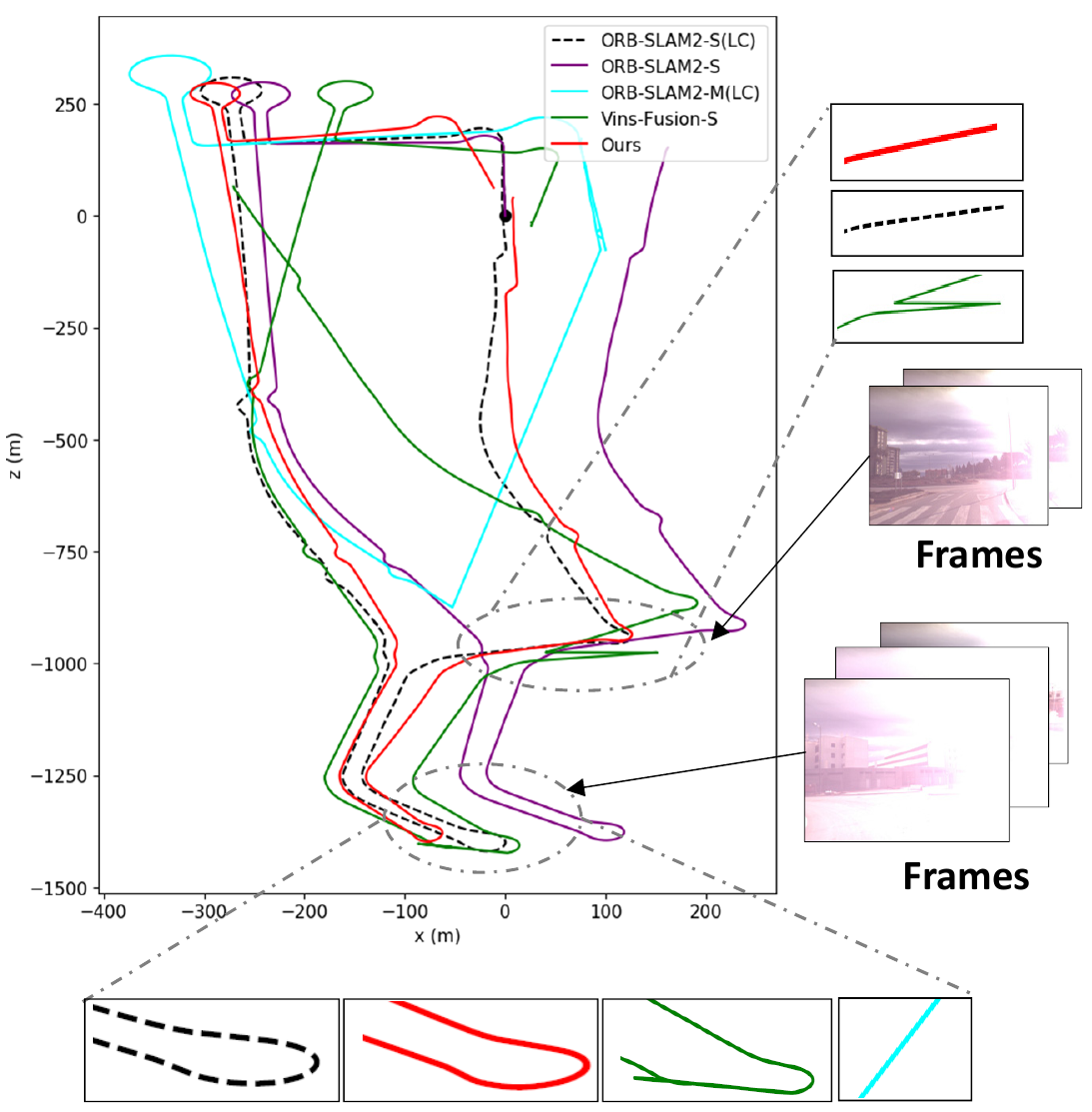}}\quad
\subfloat[Seq.14]{\includegraphics[width = 0.38\columnwidth, height = 54mm]{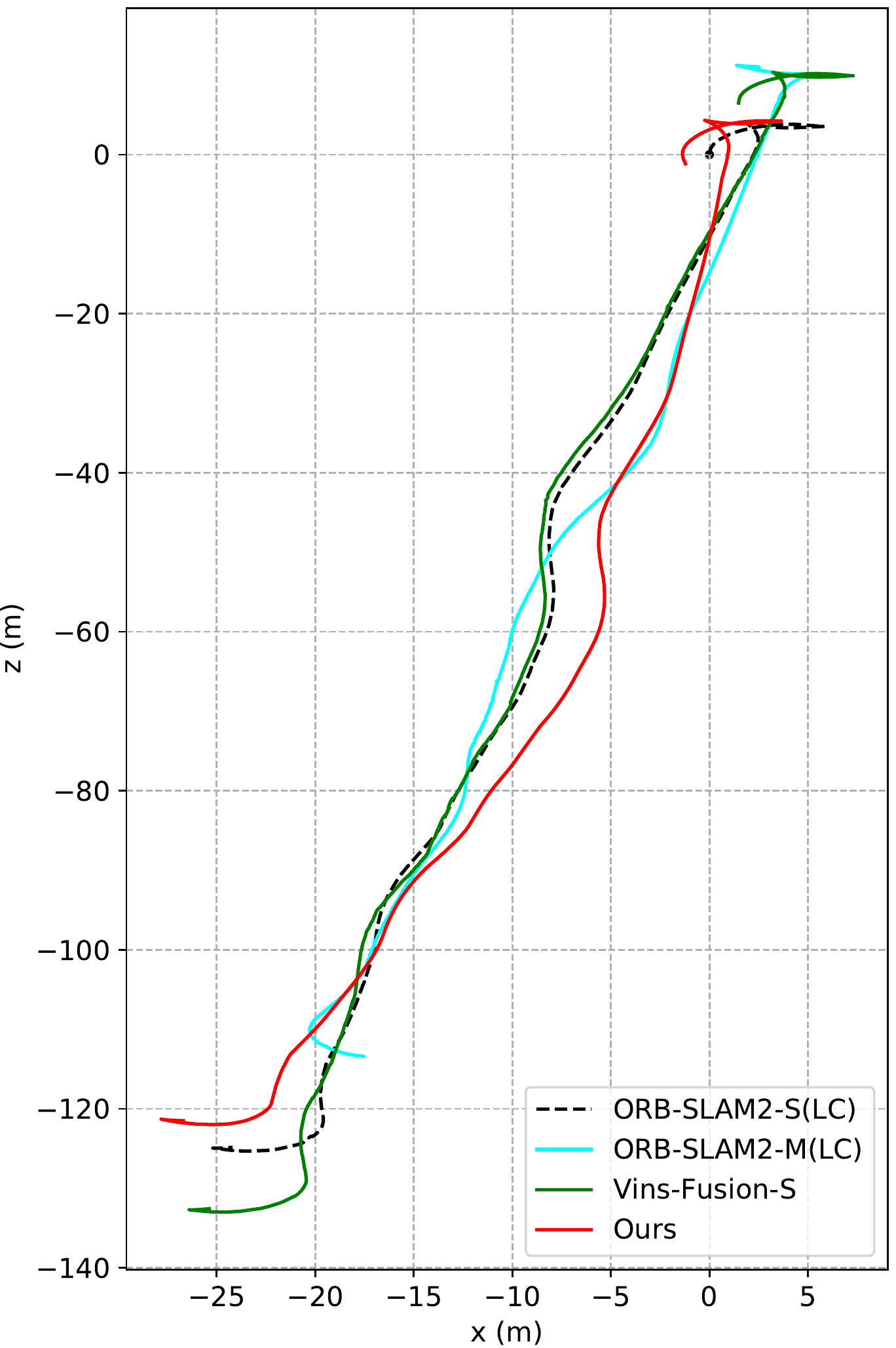}}\
\caption{Trajectories of Seq.08, 14 on Malaga dataset. ORB-SLAM2-Stereo is taken as the baseline for comparison. The frames indicates the strong light change place during inference in Seq.08.}
\label{Fig8}
\end{figure}

\subsubsection{Malaga}
The trajectories of sequences from the Malaga dataset are shown in Fig. \ref{Fig8}. Both sequences 08 and 14 are challenge scenes where the former is a long and complex scene with around 9500 frames and the latter contains scene of moving forward and reverse. The results of our method are compared to traditional VO methods ORB-SLAM2-Monocular \cite{orbslam}, ORB-SLAM2-Stereo \cite{orbslam} and VINS-Fusion-Stereo \cite{vinsfusion}. Since there is no high precision GPS data as ground truth in Malaga dataset, ORB-SLAM2-Stereo is taken as the baseline method as did LS-VO \cite{LSVO}. Our model is pre-trained on KITTI and then fine-tuned on a few data (almost 20\%) of Malaga to adjust the scale factor. It can be seen that our method achieves comparable results with ORB-SLAM2-Stereo. ORB-SLAM2-Monocular suffers from the tracking failure due to some low texture frames, while for VINS-Fusion-Stereo which utilizes optical flow to track feature points, the strong light change cause the error as highlighted in Fig. \ref{Fig8}. However our method can still perform robustly in such challenging conditions. The results show that our method can successfully generalize to the unlabelled driving scenes with different illumination conditions and camera parameters.

\begin{table*}[t]
\caption{Comparison of single view depth estimation on KITTI Eigen split. Supervision indicates which sequence used in training, ``Monocular'' means monocular sequences and ``Stereo'' means stereo image sequences with known stereo baseline.}
\centering
\setlength{\tabcolsep}{0.5mm}{
\begin{small}
\begin{tabular}{|l|c|c|c|c|c|c|c|c|}
\hline
{Method} & {Supervision} & {Abs Rel}& {Sq Rel}&{RMSE}&{RMSE log} & {\bfseries $\delta < 1.25$}& {\bfseries $\delta < 1.25^2$}&{\bfseries $\delta < 1.25^3$ } \\
\hline
\multicolumn{2}{|c|}{} & \multicolumn{4}{|c|}{lower is better} & \multicolumn{3}{|c|}{higher is better}  \\
\hline
Zhou et al. \cite{sfmlearner} & Monocular &	0.208&	1.768	&6.856&	0.283&	0.678&	0.885&	0.957 \\
\hline
GeoNet \cite{geonet}   &Monocular	&0.155	&1.296	&5.857	&{\bfseries 0.233}&	0.793&	0.931&	{\bfseries 0.973}\\
\hline
Li et al. \cite{undeepvo} &Stereo&	0.183&	1.73&	6.57	&0.268 &	/ &	/ &	/ \\
\hline
Godard et al. \cite{monodepth}  &Stereo&	0.148&	1.344	&5.927	&0.247	&0.803	&0.922&	0.964 \\
\hline
Zhan et al. \cite{zhan}  &Stereo	&0.144	&1.391	&5.869	&0.241&	0.803&	0.928&	0.969\\
\hline
Pilzer et al. \cite{refine}   &Stereo	&0.1424	&1.2306	&5.785	&0.239&	0.795&	0.924&	{ 0.968}\\
\hline
Wong et al. \cite{bilateral}   &Stereo	&{\bfseries0.135}	&\bfseries1.157	&\bfseries5.556	&0.234 & 0.820&	\bfseries0.932&	{0.968}\\
\hline
{\bfseries Ours}	&Stereo	&{\bfseries 0.135}&	{ 1.234}	&{ 5.624}	&{\bfseries 0.233}&	{\bfseries 0.823}&	{\bfseries 0.932}&	{0.968}\\
\hline
\end{tabular}
\end{small}}
\label{table:depth}
\end{table*}

\subsection{Results of Depth Estimation  \label{Subsection:depth results}}
\begin{figure*}
\begin{center}
\includegraphics[width=0.90\linewidth, height=80mm]{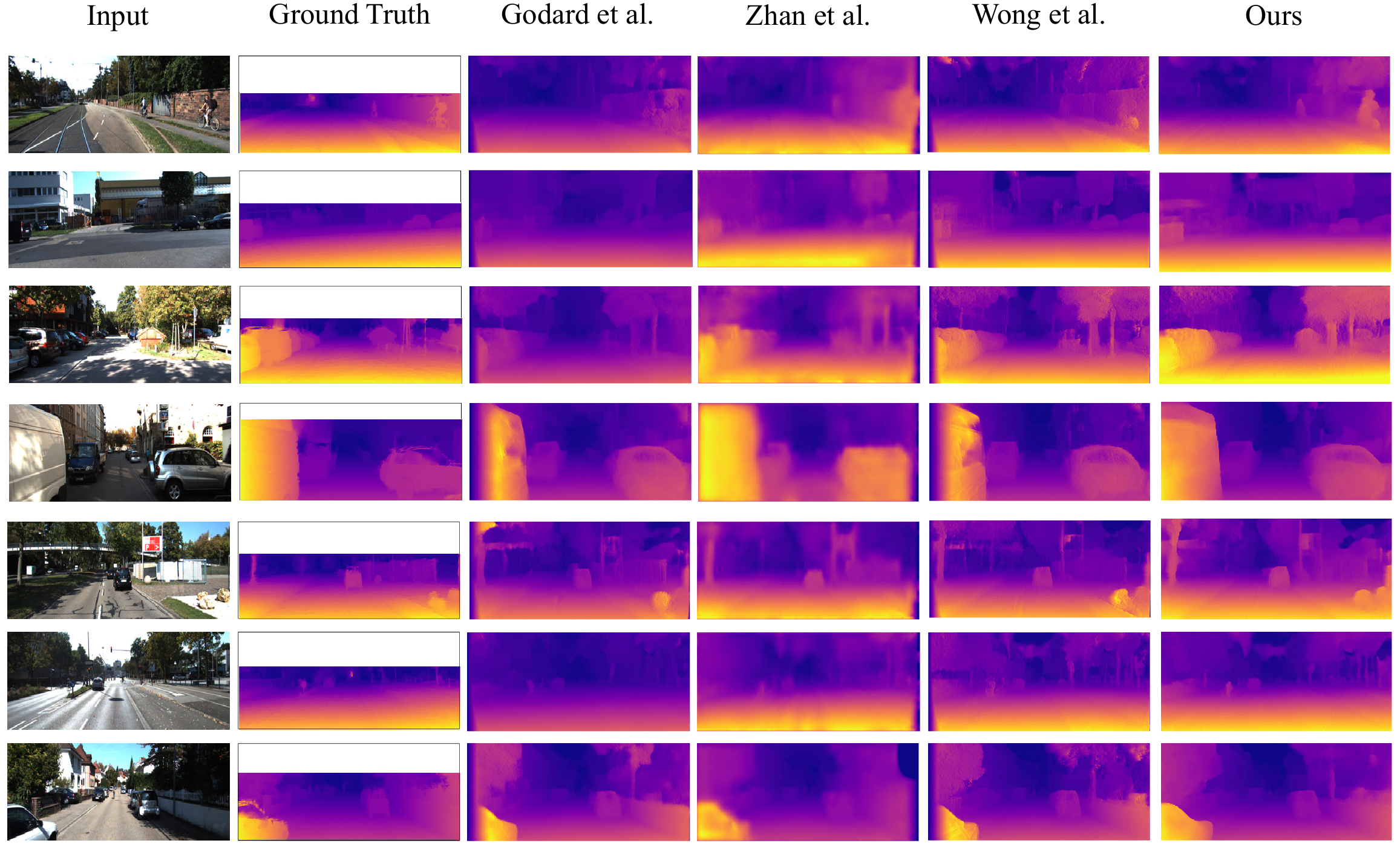}
\end{center}
   \caption{Visual comparison of the estimated depth maps on the KITTI Eigen test set. The ground truth depth is interpolated for visualization purpose.}
\label{Fig_depth}
\end{figure*}

We evaluate the depth estimation on the KITTI Eigen split \cite{eigen} using the same evaluation criterion used in monodepth \cite{monodepth}. The maximum of depth value is capped at 80 meters. The compared methods include monocular-based \cite{sfmlearner,geonet} and stereo-based \cite{monodepth,zhan,refine,bilateral}. As shown in Table \ref{table:depth}, our method achieves reasonable gains than other methods. Since our framework treats depth estimation as an auxiliary for visual odometry without special optimization, the improvement indicates that accurate camera pose estimation improves depth estimation in the proposed framework. Sample depth maps can be seen in Fig. \ref{Fig_depth}, the proposed method can recover more edge information and the boundaries of objects such as the houses and pedestrians than other compared methods.

\begin{table}[t]
\caption{Ablation study on visual odometry for different proposed modules. The evaluation data and protocol are same as Table \ref{table1}. }
\begin{center}
\begin{scriptsize}
\setlength{\tabcolsep}{0.7mm}{
\begin{tabular}{|l||c|c|c|c|}
\hline
{Method} & \multicolumn{2}{|c|}{Seq.09} & \multicolumn{2}{|c|}{Seq.10}  \\
\cline{2-5}
{ }  & \quad{$\ t_{err}$\quad}  & {\quad$r_{err}$\quad}    & {\quad$t_{err}$\quad}   & {\quad$r_{err}$\quad}\\
\hline
Baseline  & 8.76  & 3.07  & 9.88  & 3.77       \\
\hline
Baseline+FFG  &7.54	&2.30 &	6.84 &1.77  \\
\hline
Baseline+IFG  &{ 3.50}	&{ 1.74}	&{ 5.63}	&{ 1.90}\\
\hline
Baseline+F2FPE   &3.29  &	1.51&	4.89&	2.20\\
\hline
Baseline+FFG+TAPE  &{ 5.25}	&{ 2.22}	&{   5.89}	&{ 1.51}\\
\hline
Baseline+IFG+TAPE  &{ 3.66}	&{ 1.32}	&{  4.83}	&{1.58}\\
\hline
Baseline+F2FPE+TAPE   &{\bfseries 2.79}	&{\bfseries 1.07}	&{\bfseries  4.16}	&{\bfseries 1.32} \\
\hline
\end{tabular}}
\end{scriptsize}
\footnotesize
\end{center}
\label{table:ablation1}
\end{table}

\begin{table}[t]
\caption{ {Ablation study on visual odometry for different $\lambda_{pc}$. The evaluation data and protocol are same as Table \ref{table1}. }}
\begin{center}
\begin{small}
\setlength{\tabcolsep}{1mm}{
\begin{tabular}{|l||c|c|c|c|}
\hline
{$\lambda_{pc}$} & \multicolumn{2}{|c|}{Seq.09} & \multicolumn{2}{|c|}{Seq.10}  \\
\cline{2-5}
{ }  & \quad{$\ t_{err}$\quad}  & {\quad$r_{err}$\quad}    & {\quad$t_{err}$\quad}   & {\quad$r_{err}$\quad}\\
\hline
0.1  & 3.40  & 1.21  & 4.98  & 1.55       \\
\hline
0.2  &2.98	&1.55 &	4.79 & 1.77  \\
\hline
0.5  &3.10 & 1.12 &  4.21	& 1.90\\
\hline
1.0   &{\bfseries 2.79}	&{\bfseries 1.07}	&{\bfseries  4.16}	&{\bfseries 1.32} \\
\hline
2.0  &{ 2.94}	&{ 1.44}	&{  4.15}	&{ 1.43}\\
\hline
\end{tabular}}
\end{small}
\footnotesize
\end{center}
\label{table:ablation2}
\end{table}

\subsection{Ablation studies  \label{Ablation studies}}

In order to demonstrate the impact of each proposed component, ablation studies are conducted on the KITTI Odometry dataset. The baseline method is similar to Depth-VO-Feat \cite{zhan} which takes stereo image pairs as input, predicting the relative camera poses and depth (contains DepthNet, Feature Encoder (FE) and Pose Estimator (PE) in our framework). As shown in Table \ref{table:ablation1}, benefiting from the motion relation provided by optical flow, both IFG and FFG can improve the results of baseline. IFG contributes significantly since it is pre-trained firstly and can produce more accurate optical flow. In addition, TAPE can improve the performance of different baseline methods (with IFG, FFG and F2FPE respectively) with reasonable gains through the proposed pose consistency loss that encodes the geometry and temporal information from multiple images.  Table \ref{table:ablation2} reports the different weights of the $\lambda_{pc}$ used in our framework. As shown in Table \ref{table:ablation2}, the performance of the visual odometry increase when the value of weight from 0.1 to 2, and the $\lambda_{pc}=1$ produces the best performance.

\begin{figure}[t]
\begin{center}
\includegraphics[width=0.85\linewidth, height=45mm]{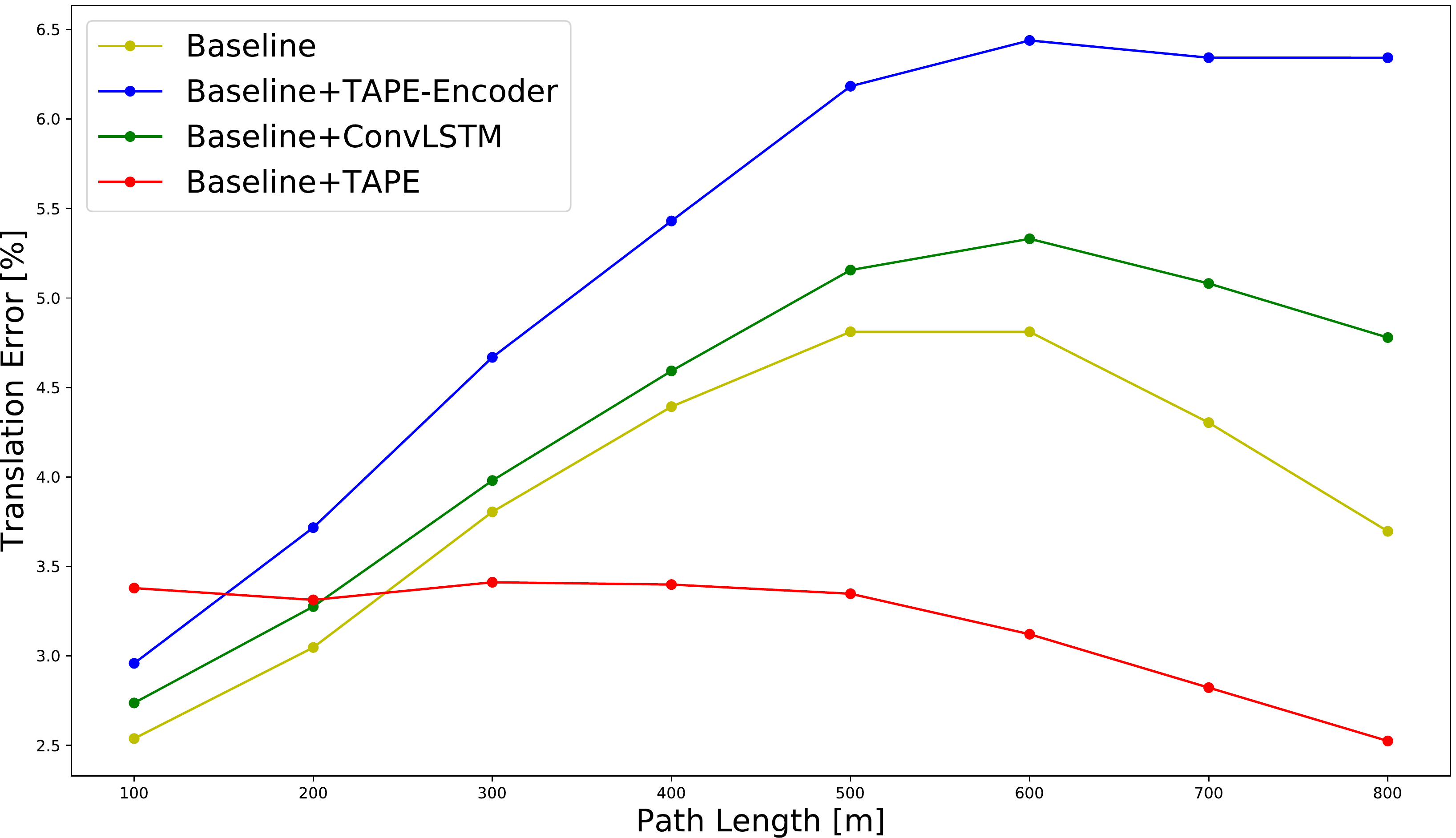}
\end{center}
\caption{Ablation study on average translation error for different module used in local temporal window. The model is evaluated on sequence 09 and 10 of odometry dataset.}
\label{Fig_ablation}
\end{figure}

 Fig. \ref{Fig_ablation} shows the results of using different modules to process input DF-Groups on a local window, where the average translation error against different path lengths is taken as the evaluation metric. It can be seen that all the modules suffer from error accumulation except TAPE. TAPE-Encoder is ResNet-50, which cannot deal with temporal information, resulting in heavy error. Although historical information can be integrated by ConvLSTM, the error increases slightly. This may mainly because the temporal window is set to a short period and the error spreads through recurrent units. Because of the self-attention mechanism, the proposed TAPE explicitly builds the contextual model in a local temporal window and effectively reduces error accumulation.

\begin{figure}[t]
\begin{center}
\includegraphics[width=0.99\linewidth, height=40mm]{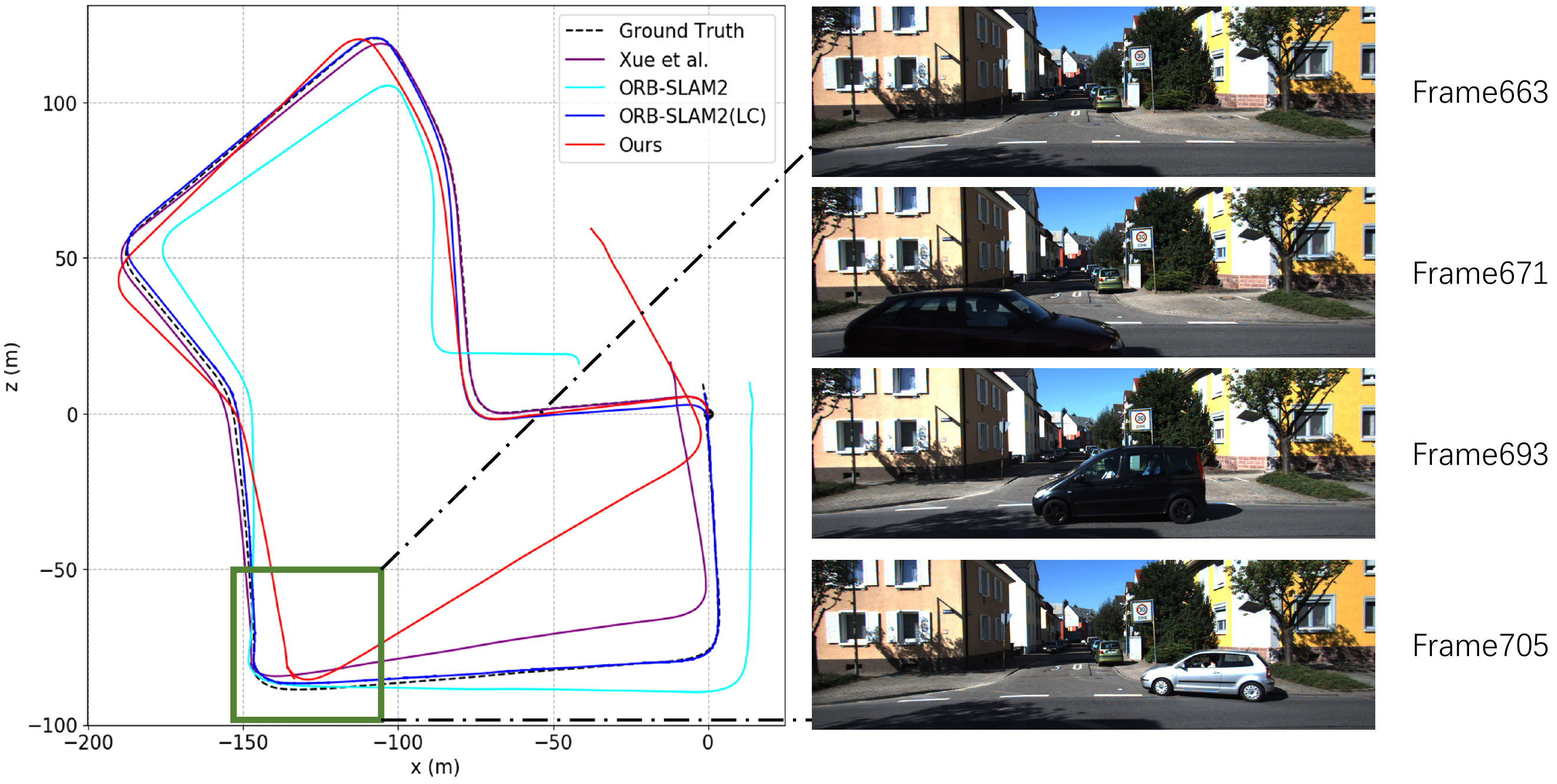}
\end{center}
\caption{ The trajectory of the Seq.07 on Odometry split, and the green box indicates the static frames.}
\label{Fig_falure}
\end{figure}

\section{Discussion and Conclusion \label{Section:Conclusion}}
In this paper, we present a novel unsupervised framework for visual odometry. Two branches called TAPE and F2FPE are proposed for camera pose estimation from multiple and pairwise frames respectively. Benefitting from the proposed pose consistency constraint, our framework can effectively leverage the information from both estimators to produce accurate camera poses and reduce error accumulation significantly. Experiments have shown that the proposed framework achieves state-of-the-art results among the unsupervised methods. The results are comparable to those obtained by supervised and traditional methods. Moreover, our method can perform robustly in low texture and changing scene lighting, where the traditional methods often fail. 

  Our approach still has limitations. As described in Sec. \ref{Subsubsection:KITTI}, our method suffers from rotation error in Seq. 07. Fig. \ref{Fig_falure} shows the trajectory of Seq. 07, and the green box shows the static frames which cause the error. In addition, the performance of the depth estimation is limited. For future work, we would firstly reduce the error caused by static scenes and then consider utilizing cross-attention structure \cite{casnet} to enhance inter-modality geometry dependence for depth, optical flow and camera poses in our framework. Finally, we would like to extend the unsupervised VO to a full unsupervised SLAM system which contains additional optimization and mapping modules.

\textbf{Acknowledgements} This study was funded by National Natural Science Foundation of China (grant numbers 61906173,61822701).
\section*{Compliance with ethical standards}
\textbf{Conflict of interest} The authors declare that they have no conflict of interest.

\end{document}